\documentclass[conference]{IEEEtran}
\IEEEoverridecommandlockouts                              
\usepackage[bookmarks=false]{hyperref}
\usepackage{blindtext, graphicx}
\usepackage{cite}
\usepackage{graphicx}
\usepackage{fixltx2e}
\usepackage{csquotes}
\usepackage{caption}
\usepackage{subcaption}
\usepackage{amssymb}
\usepackage{amsmath}
\usepackage{siunitx}
\usepackage[font={small}]{caption}	
\usepackage{fancyhdr}
\usepackage{todonotes}
\usepackage{booktabs,tabularx,ragged2e,url}
\usepackage{mathrsfs}
\DeclareMathAlphabet{\mathpzc}{OT1}{pzc}{m}{it}
\usepackage[ruled]{algorithm2e}

\usepackage{comment}

\graphicspath{{Figures/}}
\newcommand{\fig}[1]{Fig.~\ref{#1}}

\usepackage{enumitem,booktabs}
\usepackage{xcolor}
\usepackage[referable]{threeparttablex}
\renewlist{tablenotes}{enumerate}{1}
\makeatletter
\setlist[tablenotes]{label=\tnote{\alph*},ref=\alph*,itemsep=\z@,topsep=\z@skip,partopsep=\z@skip,parsep=\z@,itemindent=\z@,labelsep=.2em,leftmargin=*,align=left,before={
\footnotesize}}
\makeatother

\DeclareCaptionLabelSeparator{periodspace}{.\quad}
\usepackage{watermark}

\begin{document}
\title{Genesis: A Spiking Neuromorphic Accelerator With On-chip Continual Learning}
%

\author{Vedant Karia, Abdullah Zyarah,  and Dhireesha Kudithipudi

\\ Neuromorphic AI Lab, University of Texas at San Antonio, TX, USA
\thanks{This effort is partially supported by NSF EFRI BRAID Award \#2317706, NSF NAIAD Award \#2332744 and Air Force Research Laboratory under agreement number FA8750-20-2-1003 through BAA FA8750-19-S-7010. The views and conclusions contained herein are those of the authors and should not be interpreted as necessarily representing the official policies or endorsements, either expressed or implied, of NSF, Air Force Research Laboratory or the U.S. Government}}

\maketitle
\begin{abstract}


Continual learning, the ability to acquire and transfer knowledge through a model's lifetime, is critical for artificial agents that interact in real-world environments. Biological brains inherently demonstrate these capabilities while operating within limited energy and resource budgets. Achieving continual learning capability in artificial systems considerably increases memory and computational demands, and even more so when deploying on platforms with limited resources. In this work, Genesis, a spiking continual learning accelerator, is proposed to address this gap. The architecture supports neurally inspired mechanisms, such as activity-dependent metaplasticity, to alleviate catastrophic forgetting. 
It integrates low-precision continual learning parameters and employs a custom data movement strategy to accommodate the sparsely distributed spikes. Furthermore, the architecture features a memory mapping technique that places metaplasticity parameters and synaptic weights in a single address location for faster memory access. Results show that the mean classification accuracy for Genesis is 74.6\% on a task-agnostic split-MNIST benchmark with power consumption of 17.08 mW in a 65nm technology node.

\end{abstract}




\section{Introduction}
Machine learning models are promising for a wide range of tasks that have static data distributions. In contrast, when trained on non-stationary data distributions or sequential tasks, these models fit their parameters to the latest distribution while suffering catastrophic forgetting of previously learned information~\cite{french1999catastrophic}. The ability to learn from non-stationary data streams is crucial for intelligent agents interacting with real-world environments where the data stream does not satisfy the I.I.D. . assumption. This has motivated research efforts towards continual learning (CL), in which an agent learns and transfers knowledge across sequential tasks while reducing catastrophic interference~\cite{hadsell2020embracing, parisi2019continual}. However, most of the efforts incur high computational cost and memory overhead~\cite{de2021continual, wang2024comprehensive}.

Biological brains, on the other hand, are inherently able to learn continually while operating within minimal resource and energy budgets~\cite{kudithipudi2022biological}. Prior studies have proposed several mechanisms to explain this ability, and researchers have attempted to emulate them in artificial systems by introducing new learning methodologies, models, and architectures~\cite{furber2016brain, strukov2019building, li2024brain} inspired by the brain. Among these approaches, spiking networks, with their binary activations and event-driven nature, offer a promising low-cost, low-power solution to learning~\cite{survey_neuromorphic_CL}. Combining them with other neuro-inspired mechanisms such as metaplasticity~\cite{abraham_1996_Metaplasticity} would enable efficient continual learning capabilities in complex task-agnostic scenarios.



Although spike-based continual learning solutions are promising to improve the efficiency of continual learning~\cite{laborieux2021synaptic, soures2021tacos, jedlicka2022contributions, zohora2021metaplasticnet}, there still exist significant challenges for designing hardware architectures capable of learning continually~\cite{kudithipudi2023design}.
The main challenges can be divided into two categories. First, CL mechanisms introduce additional computations, which inevitably increase data movement and communication overhead. Data movement requires orders of magnitude higher energy than computations \cite{dally2022model}, and severely strains the energy budget while considering edge platforms. Second, CL mechanisms often require additional parameters, which are challenging to store in the limited memory available in edge devices. For instance, activity-dependent metaplasticity can require an additional parameter per network weight, doubling the memory required to store parameters. A common optimization technique to minimize the memory overhead is quantization. However, most approaches consider quantization only during inference while relying on high-precision parameters for training. This can be attributed to the loss of information when network parameters are quantized during training, leading to a catastrophic degradation in performance \cite{putra2021q}. These challenges are persistent, if not worse, in continual learning settings where the model trains on changing data distributions. The above-mentioned aspects render achieving continual learning within edge constraints extremely challenging, and spike-based hardware architectures for continual learning on edge remain under explored.

%

\begin{figure*}[!ht]
\centering
\includegraphics[width=0.74\textwidth, height=0.45\textwidth]{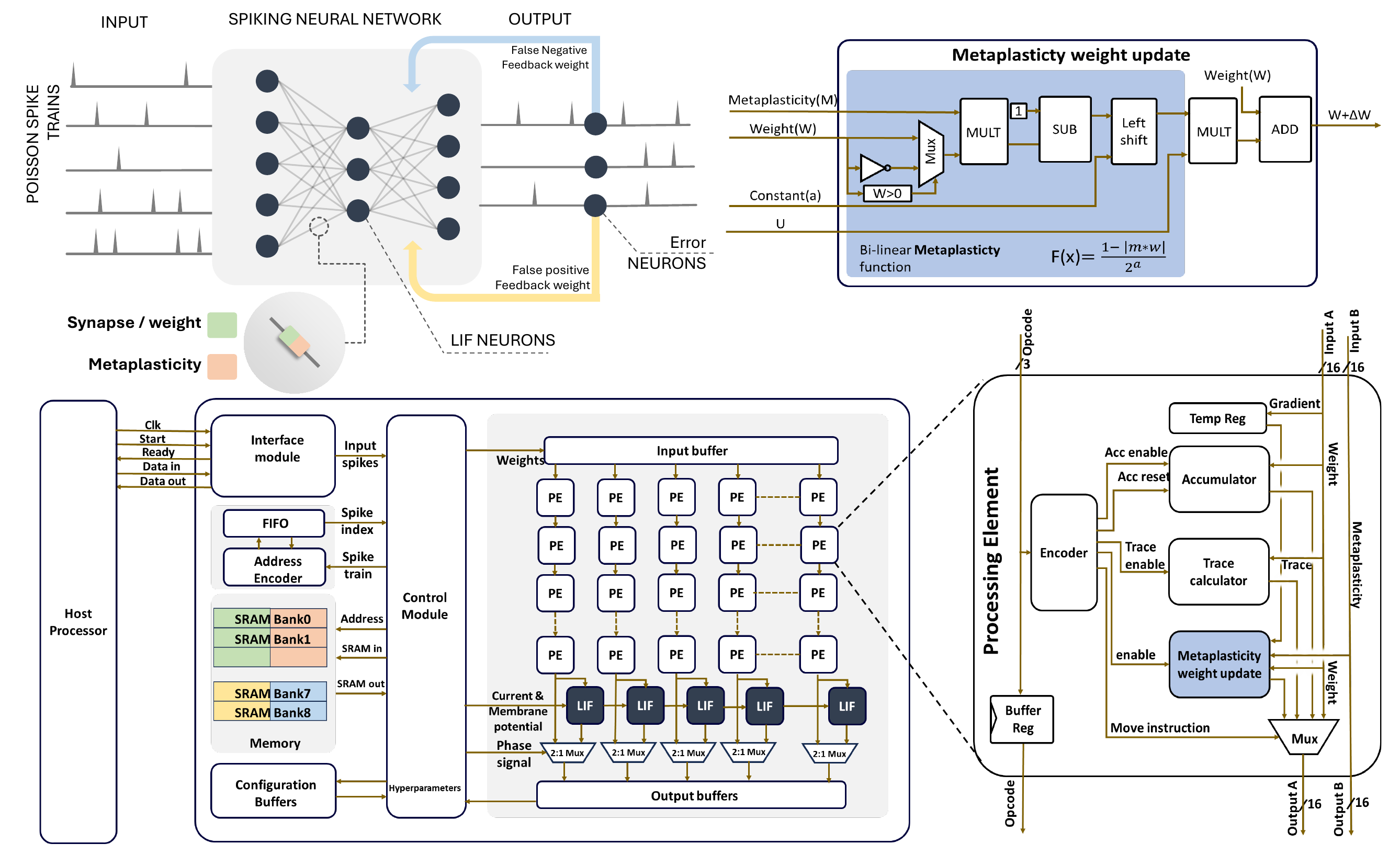}
      \caption{ Architecture overview of the Genesis, with  8x8 incorporates processing elements (PE) arranged in mesh topology. Each PE in the systolic array is modeled to perform 8 instructions, 3 compute instructions (accumulate, reset accumulator, metaplasticity weight update) and 5 data movement instructions (Load temp, load accumulator, move input, move weight, move accum). Each PE is equipped with a trace calculator to monitor the activity of neuron, an accumulator for weighted sum of spikes, and a metaplasticity weight update unit that modulates the plasticity of synaptic connections.}
      \label{fig:Systemarch}   
\end{figure*}


To address these challenges, we introduce Genesis, the first spiking continual learning accelerator. Genesis integrates optimizations across the algorithmic and hardware levels to enable efficient continual learning with SNNs under sub $20mW$ power. It supports surrogate gradient-based training enhanced with activity-dependent metaplasticity, leveraging neuron-local information to reduce communication overhead. As part of the hardware design, we develop custom processing elements optimized for metaplasticity and activity trace computation. Moreover, these elements utilize a custom Address-Event Representation (AER) protocol to minimize data movement during training. As a result of the neuro-inspired learning mechanisms and optimizations, we observe that Genesis alleviates catastrophic forgetting while halving memory overhead using quntization and achieving power consumption and throughput of 17.08 mW and 640 MOPS. We perform additional analysis to demonstrate the benefits of the proposed optimizations in comparison to other dataflow schemes, PE configurations, and bit precision. 


The main contributions of Genesis work include:
\begin{enumerate}

    \item A spiking neuromorphic CMOS ASIC for on-chip online task-agnostic continual learning.
    
    \item  A metaplasticity enhanced processing element that continually learns in domain-incremental learning scenarios with quantized (16-bit) parameters.

    \item An AER encoding with a custom data movement strategy for sparsely distributed spikes.
    
    \item A memory mapping scheme that co-locates metaplasticity parameters with weights to reduce the memory fetches by half.
    

\end{enumerate}

\section{Background and Related Works}
In this section, we detail the widely adopted methods for continual learning and the emerging class of accelerator designs that support different features for such learning.

\subsection{Related work}

Continual learning is an approach in which a model is expected to learn from noisy, unpredictable, changing data distributions and continually consolidate knowledge about new information while preventing the system from catastrophically forgetting old knowledge. Several researchers have explored mechanisms inspired by the brain that enable CL models to mitigate catastrophic forgetting and can be broadly classified into (i)  loss or functional parametric adjustments, i.e., regularization~\cite{ewc, zenke2017continual}, (ii) dynamic architectures, including neurogenesis~\cite{ ebrahimi2020adversarial,rusu2016progressive}, and (iii) rehearsal or replay of previous experiences~\cite{aljundi2019grad, van2024continual}. In this work, we incorporate a regularization-based technique to address catastrophic forgetting and thereby focus primarily on comparison with other regularization mechanisms. Regularization methods aim to protect critical parameters from significant changes during new task learning. For instance, Stochastic Synapses (SS)~\cite{schug2021presynaptic} adopt probabilistic weight transmission, where weights with a high transmission probability are more strongly consolidated. Bayesian Gradient Descent (BGD)~\cite{zeno2018task} is a Bayesian-inspired continual learning method that evaluates parameter importance by considering the uncertainty in their distribution. Learning without Forgetting (LwF)~\cite{li2017learning} preserves a model's responses to previous tasks through knowledge distillation, ensuring that new learning does not interfere with prior knowledge. TACOS~\cite{soures2021tacos} introduces activity-dependent metaplasticity, where weights are consolidated based on both their magnitude and the activity level of adjacent neurons, enhancing the stability of previously learned tasks. However, all the aforementioned mechanisms to alleviate forgetting require additional memory and computations. To make CL viable for deployment on resource-constrained platforms, studies have started exploring novel hardware architectures and dataflows to accelerate learning with the additional overhead~\cite{piyasena2021accelerating, ressa2024tinycl, karia2022scolar, ravaglia2021tinyml}.


Although various accelerators have been proposed for on-device continual learning, most remain limited to FPGA prototypes~\cite{karia2022scolar, piyasena2020dynamically} and incorporating Replay based mechanisms depending on memory buffers~\cite{ressa2024tinycl, ravaglia2021tinyml, ravaglia2020memory}. Few of these replay based accelerators are ~\cite{ravaglia2021tinyml, ravaglia2020memory, ressa2024tinycl} which employs a latent replay technique on an RISCV based edge device PULP~\cite{rossi2015pulp}, making it a practical approach for edge devices by incorporating quantization where the replay buffer is compressed by reducing the precision of replay samples to 7-bits. Similarly, ~\cite{aggarwal2023chameleon} utilizes short-term and long-term memory mechanisms to reduce power consumption by minimizing off-chip memory access of replay buffer. Limited accelerators incorporate regularization mechanisms in the accelerators, such as in \cite{karia2022scolar}, leverage metaplasticity with complex synapse models in spiking neural networks to mitigate catastrophic forgetting. It incorporate methods such as dual fixed-point quantization and dyadic scaling, which have proven effective for edge computing. In this work, we adopt a similar approach. Another example, ~\cite{piyasena2021accelerating}, is an edge accelerator with continual learning (CL) capabilities, using a custom streaming linear discriminant analysis (SLDA) approach in a convolutional neural network. It employs optimizations such as on-chip data reuse and memory organization to make it suitable for edge applications. Moreover, ~\cite{piyasena2020dynamically}, employ dynamically growing networks like Self-Organizing Neural Networks (SONN). However, these approaches face challenges related to increased memory requirements and compute limitations on edge devices.

\begin{figure*}\captionsetup[subfigure]{font=footnotesize}
     \centering
     
     \begin{subfigure}[b]{0.3\textwidth}
         \centering
         \includegraphics[width=0.99\linewidth]{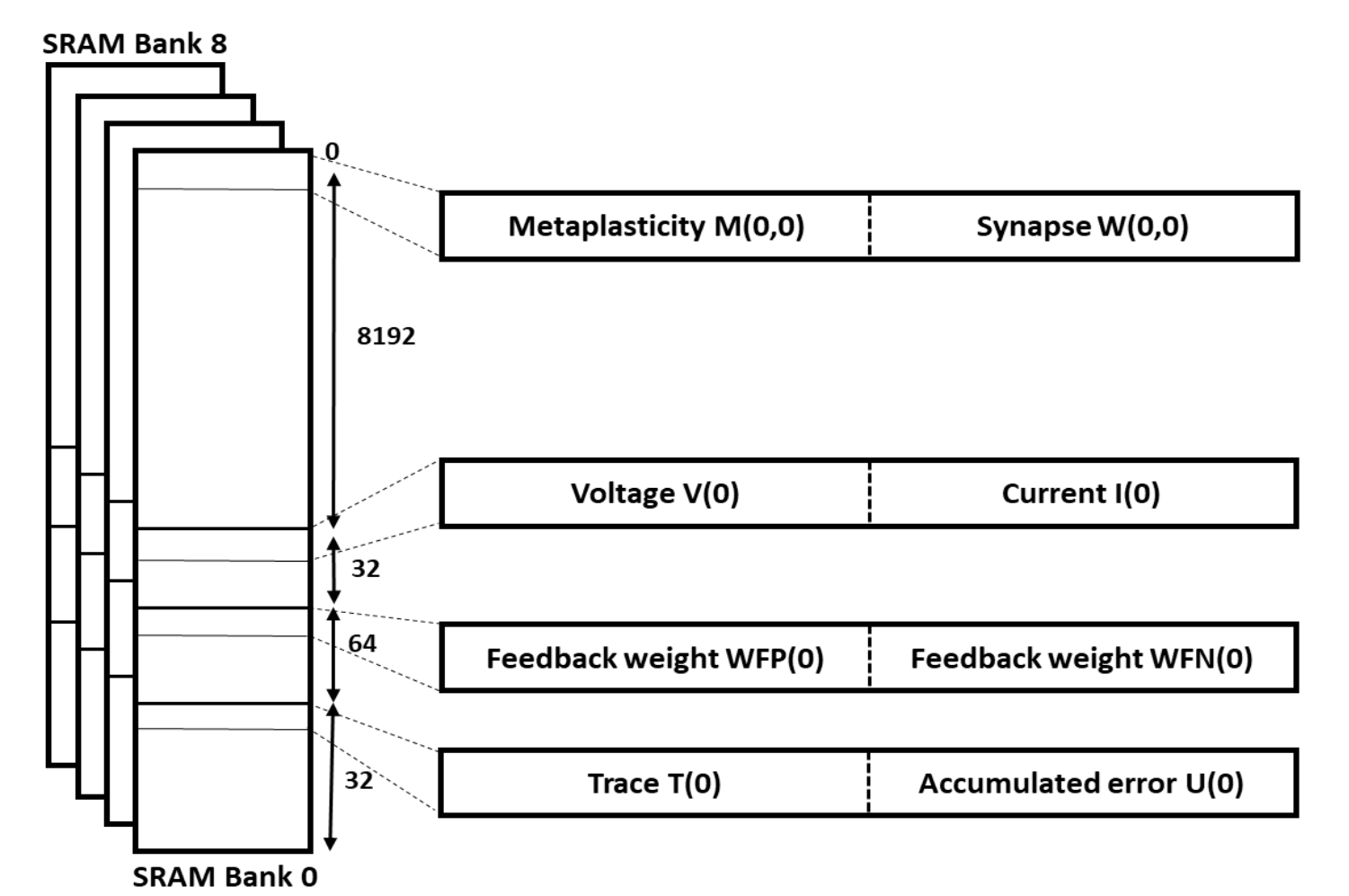}
         \caption{}
         \label{fig:Mem_ord}
     \end{subfigure}
     \begin{subfigure}[b]{0.55\textwidth}
         \centering
         \includegraphics[width=0.99\linewidth]{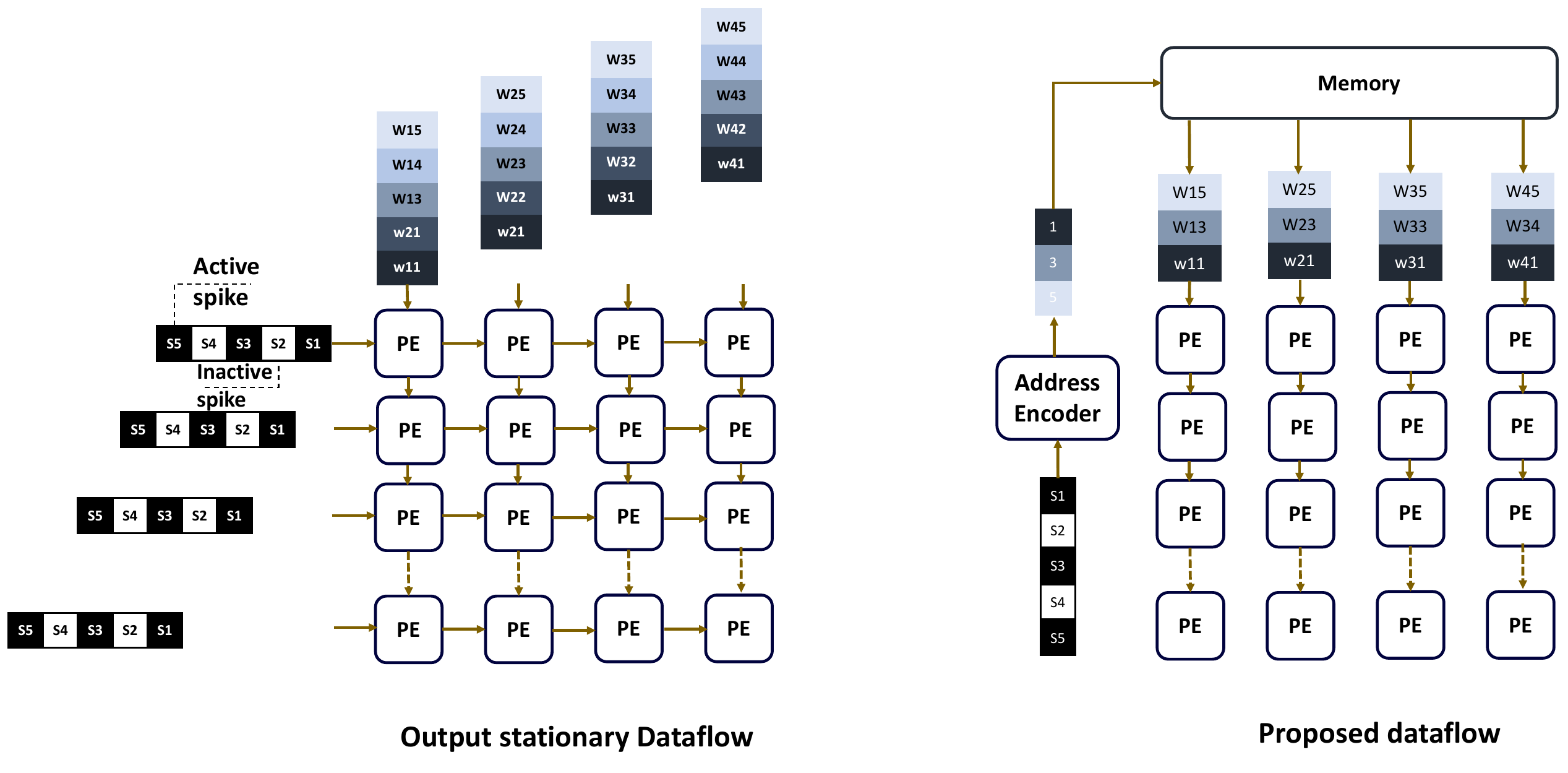}
         \label{fig:datamovement}
         \caption{}
     \end{subfigure}
        \caption{(a) Memory mapping strategy of the network parameters with metaplasticity and synaptic weights coupled in a single memory address within the interleaved SRAM bank, for fast read/write cycles. (b) demonstrates the distinction between the widely-used output stationary dataflow and the newly proposed dataflow, which incorporates an address encoder to encode the index of the active neuron and relocate the parameters linked to that neuron.}
        \label{fig:}
\end{figure*}

\subsection{Spiking Network for Continual Learning}


The SNN model is a two-layer feedforward network with neuro-inspired mechanisms such as metaplasticity and consolidation. In addition to being resource efficient, these mechanisms intrinsically function with spiking neurons with integrated potentials and multiple compartments. Metaplasticity tracks the change in synaptic strength to measure the synapse's importance. Depending on importance, updates to the corresponding synapse are regularized by an bi-linear metaplasticity function. 





\begin{equation}\label{currenteq1} I(t+1) = I(t) + 2^a * \left(\sum_{j=1}^{N} w_j S_j(t) - I(t)\right) \end{equation}
\begin{equation}\label{Voltageeq1} V(t+1) = V(t) + 2^b * (V_{rest}-V(t)) + 2^c * I(t) \end{equation}

In the hidden layer and the output layer, error currents are calculated by comparing the output neuron spikes with the label neuron spikes. These error currents, along with their inverses, are passed to two error-encoding neurons that generate false positive and false negative error spikes, respectively. The weight updates for the output layer are directly computed using these error spikes, while error signals are propagated through fixed random weights to the hidden layer neurons.
Each neuron has a dendritic compartment, denoted as $U$, which integrates incoming error spikes $E(t)$ as described in equation ~\ref{dendritic}. Weight updates occur whenever a presynaptic spike is detected, with the update rule governed by the learning rate $\eta$, the presynaptic spike binary variable $S_j$, and the postsynaptic error signal $U_i$. Additionally, $\Theta (I_i(t))$ is a boxcar function that is active only when the postsynaptic current ($I(t)$) lies within the limits ($I_{min},I_{max}$)

\begin{equation}\label{dendritic}
    U(t + 1) = U(t) + {\frac{\Delta{t}}{\tau_{mem}}}(E(t)R)
\end{equation}

\begin{equation}\label{weight_update}
    w_{i,j}(t+1) = w_{i,j}(t) - \eta S_j(t)U_i(t)\Theta(I_i(t))
\end{equation}

The network uses a set of leaky integrate-and-fire neurons, that operate by accumulating weighted inputs over time. However, instead of immediately applying an activation function, the input is integrated with a decay factor, as modeled by equations ~\ref{currenteq1} and ~\ref{Voltageeq1} where I is the input current of the neuron at time t, V is the membrane potential and $a, b, c \in \mathbb{Z}^{+}$ are configurable constants to model the neuron. Once the integrated value surpasses a predefined threshold, the neuron generates a voltage spike.

 To enable continual learning, to preserve the old knowledge we penalize the synapse by regulating the plasticity of synapse. The plasticity of synapse wholly depends on the synaptic strength and the metaplastic strength. The plasticity of the synapse is calculated based on the bi-linear function $f(m,w)$ shown in the equation ~\ref{meta_func}, where m is the metaplasticity parameter, w is the synaptic strength and $d$ is a user configurable parameter which determines the area under the curve(AUC) of the function. The AUC is direcly proportional to the plasticity.    

\begin{equation}\label{meta_func}
    f(w,m) = 1- {\frac{|m * w|}{ 2^{d}}}
\end{equation}

To regularize the plasticity, the weight update equation~\ref{weight_update} is modified by multiplying the $f(w.m)$ is multiplied by the $\eta$ learning rate. This function reduces the plasticity of the synapse to preserve the old knowledge.

\section{Genesis Hardware architecture}

The high-level architecture of Genesis, shown in \fig{fig:Systemarch}, consists of a systolic array of interconnected processing elements (PEs) arranged in a mesh topology with 8 columns and 8 rows, an address encoder, and an interface module. The PEs are responsible for computing the weighted sum of spikes that are transmitted from the host processor, computing the traces of neurons and the importance of each synapse using the metaplasticity parameter, and modulating the synaptic strength. 
The address encoder is responsible for encoding the pulse indices in the spike train and storing them in the FIFO buffers, 
whereas the interface module bridges Genesis with a general-purpose host processor for initializing and configuring the accelerator via a 16-bit parallel full-duplex data bus.



To ensure efficiency, the accelerator runs in four distinct phases: initialization, forward pass, backward pass, and synapse update. The initialization phase is utilized only once when the accelerator is run for the first time. During this phase, all synaptic connections are randomly initialized, and the network configuration settings are set via the host processor.
After initialization, the forward pass begins, in which the host processor transfers the input spike trains to Genesis. The spike trains are then routed to the address encoder to encode the pulse indices and store them in the FIFO buffer. The control unit reads the FIFO indices to access the synaptic weights connected to the active neurons. After this step, the control unit transfers the synaptic weights to the PEs in a pipelined fashion and then processes the weighted sum of spikes across 64 PEs arranged in an 8x8 2D configuration. Each PE is equipped with a 16-bit accumulator and a 16-bit reconfigurable metaplasticity weight update unit. Additionally, each PE accumulates and holds the partial output for a specific neuron. The accumulated weighted sum is then pipelined into configurable leaky integrate-and-fire (LIF) neuron modules that are connected to the last row of PEs, where the neuron’s membrane potential and output spike are computed. The constants in the network are scaled dyadically, allowing the majority of the multipliers to be replaced with shift operations, as shown in equations \ref{currenteq1} and \ref{Voltageeq1}. The computed output spikes and respective membrane potential from the LIF neurons are then transferred to the memory. Similarly, the output neuron spikes are computed and stored in the memory.  

\begin{figure*}[!ht]
\centering
\includegraphics[width=0.9\linewidth]{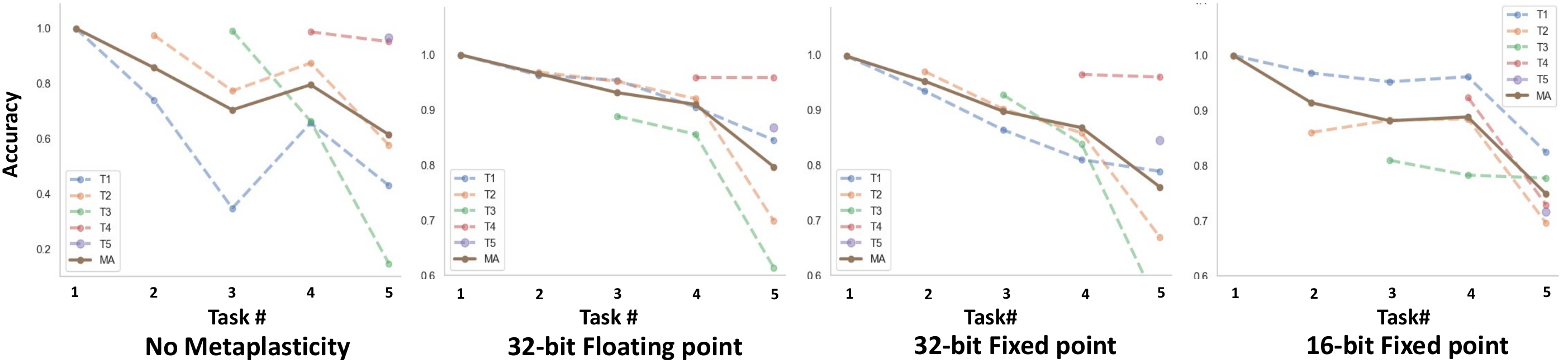}
      \caption{\textbf{Quantization of model parameters to half precision degrades mean accuracy (MA) by 5\%}. The SNN model without metaplasticty  achieves $\sim$61\% mean classification accuracy while the metaplasticity-based model realized on 32-bit floating point boosts the performance by 18\%. Quantizing the parameters from floating point to fixed point degrades performance by 3\%. Moreover, on reducing the precision to 16 bits, minimal degradation (2\%) of accuracy is noticed}
      \label{fig:Acc_results}   
\end{figure*}

     

\begin{table*}[!h]
\begin{center}
\caption{\textbf{Detailed study of existing accelerators with on-device continual learning capabilities is shown}. Majority of the existing CL accelerators are prototyped on a FPGA or realized on existing edge devices by extending the off-chip memory for replay buffer.}
\begin{tabular}{@{}lcccccccc@{}}
\toprule
  &  \textbf{Mechanism} & \textbf{Scenario} & \textbf{Technology} & \textbf{On-chip memory(MB)} & \textbf{Bit-preicion} & \textbf{Power} & \textbf{Latency}\\
\midrule

  SLDA ~\cite{piyasena2021accelerating} & Regularization & Class-IL & FPGA & 1.25& 32-bit-FXP/16-bit FXP  & 4.81 W & \textbf{5.88ms} \\
  MLA*~\cite{ravaglia2020memory} & Reply & Class-IL & ASIC(22nm) & 1.3 & 32-bit FP  & 70 mW & 502 ms \\
  SONN~\cite{piyasena2020dynamically}\ & Dynamic arch & Class-IL & FPGA & 1.66 & 16-bit FXP/ 8-bit FXP  & -- & -- \\
  TinyCL~\cite{ressa2024tinycl} & Replay & Class-IL & ASIC(65nm) & 1.3 & 16-bit FXP  & 86 mW & -- \\
  TinyML*~\cite{ravaglia2021tinyml} & Reply & Class-IL & ASIC(22nm) & 1.3 & 32-bit FP  & 70 mW & 502 ms \\
  SCOLAR~\cite{karia2022scolar} & Regularization & Domain-IL & FPGA & \textbf{0.2} & \textbf{8-bit FXP}  & 21.1 mW & 30 ms \\
  \textbf{Genesis} (our work) & Regularization & Domain-IL & ASIC(65nm) & \textbf{0.5} & 16-bit FXP/ \textbf{8-bit FXP}  & \textbf{17.08 mW} & 10 ms \\

\bottomrule
\end{tabular}
 \begin{tablenotes}
       \item [*] System is built on PULP~\cite{rossi2015pulp} a RISC-V-based IoT device.
\end{tablenotes}
\label{tab:Split_MNIST}
\end{center}
\vspace{-6mm}

\end{table*}
During the backward pass phase, each PE integrates false-positive and false-negative error neurons. Once the processing elements compute the error for each output neuron, the spikes of these error neurons are address-encoded, and the weighted sum of the error neurons is transferred to the LIF neurons to compute gradients, as shown in equation \ref{dendritic}.

In the synapse update phase, whenever a presynaptic neuron fires, the weights of the synaptic connections bridging postsynaptic and presynaptic neurons are updated. This weight update process follows similar procedure to forward pass while the weights associated to active neurons are transferred to PEs to accumulate, here the weights are transferred to be updated. The weights connected to the active presynaptic neurons are accessed from memory and transferred to the PEs. Before this step, the $U$ values associated with each neuron are stored in the PE, which is essential for model training. When the weight and its associated metaplasticity parameter reach the respective PE, they are processed through a configurable bilinear function to compute the importance of the synapse, thereby regularizing the weight update. The output of this bilinear function determines the synapse’s relevance to previously learned information and is multiplied by the corresponding $U$ value, as shown in Figure \ref{fig:Systemarch}.

After the model is trained on each sample, the metaplasticity parameters are updated in the PEs based on the trace collected during the forward pass phase. If the postsynaptic trace crosses a threshold, the metaplasticity parameter is strengthened; conversely, when the presynaptic trace crosses a threshold, the metaplasticity parameter is weakened.

\section{Design Methodology}

\subsection{Low-precision Continual Learning Parameters}
Metaplasticity-based approaches, which introduce additional parameters to manage synaptic dynamics, further increase memory demands, often doubling the model's memory footprint. To mitigate this overhead, the SNN parameters  are converted from full-precision floating-point to a 16-bit fixed-point representation. This optimization effectively reduces the memory size requirements by $2\times$, thereby enhancing both area and power efficiency. Specifically, it achieves a  $58.4\%$ reduction in area and improves the power efficiency by approximately $52.2\%$ in Genesis architectures.
These combined benefits make quantization a highly effective solution for optimizing continual learning models that take advantage of metaplasticity.


In addition to using 16-bit fixed-point precision (FXP), we explore low precision to calculate the trace parameter, which defines the spiking activity of individual neurons. The trace ($X_{tr}$), where $\frac{d}{dt}X_{tr} = -\frac{X_{tr}}{\tau_{tr}} + S$, is calculated using leaky integrator with leakage $\tau_{tr}$ and neuron output spikes ($S$). The trace of each neuron is generated using an 8-bit fixed-point format rather than a 16-bit format due to the minimal dynamic range required (the maximum value the trace achieves by integrating spikes is 100). This results in a 42.79\% improvement in energy efficiency.


\subsection{Data flow}

Widely used topologies for designing systolic arrays—such as output stationary, input stationary, and weight stationary—are highly efficient for CNNs. However, these techniques do not provide considerable advantages when mapped for SNNs. This is because SNNs typically feature sparse binary input activations, making traditional data movement strategies inefficient. In SNNs, several neurons may be inactive at any given time, which results in unnecessary data movement for both the inputs and the weights associated with these inactive neurons.

To address this issue, we propose a novel address   encoding technique. This technique identifies active neuron indices and extracts the weights associated with these active neurons from the memory banks, as illustrated in Figure 2(c). The relevant weights are then moved directly into the corresponding PEs for accumulation, eliminating the need to transfer the inputs or weights associated with inactive neurons. This approach significantly reduces data movement, improving the efficiency of spike processing in systolic array-based designs.

\subsection{Memory Organization \& Mapping}
Genesis consists of 8 PEs in each row that independently handle synaptic accumulation and synaptic consolidation. Using a single memory bank limits the system throughput due to bandwidth constraints. To overcome this, memory is distributed across eight banks with the low-order interleaving scheme, where the consecutive memory addresses are in different memory banks for faster read/write access. This distributed memory organization increases the memory bandwidth by a factor of $8$.

During the weight update phase, updating each synapse requires accessing metaplasticity parameters to compute synaptic plasticity, resulting in two memory accesses per synapse update. To optimize this process, the metaplasticity parameters and the synapse data are stored at the same memory address, with 16 bits allocated for each data point. This design enables both parameters to be retrieved in a single cycle, reducing the total number of memory read/write cycles. By co-locating parameters, the system achieves a $\sim$20\% reduction in training latency.

\section{Results and Analysis}
\subsection{Continual learning performance}
We evaluate the performance of the Genesis accelerator on the Split-MNIST continual learning task in the domain-incremental learning (Domain-IL) scenario~\cite{vandeven_2019_Three}. In this scenario, the output neurons are shared among multiple tasks, a challenging setting to study the effect of catastrophic forgetting due to task interference. Training is performed in a task-agnostic schema where the network is not provided with explicit information of the change in task, making a suitable scenario for online learning. The tasks are presented sequentially, each involving two classes of MNIST digits. After completing the training on all tasks, we assess the network's continual learning capability by calculating its mean accuracy across all tasks.

\begin{table}[!h]
\setlength\tabcolsep{3 pt}
\begin{center}
\caption{The mean accuracy of regularization-based continual learning models across all tasks and the memory size required to store the parameters to mitigate catastrophic forgetting. 
}
\begin{tabular}{@{}lcc@{}}

\toprule
 \textbf{CL Methods}  &  \textbf{Mean Accuracy} & \textbf{Memory Size}\\ 
 
\midrule
Baseline & $61.43\%$ & 628 kB \\
LwF~\cite{li2017learning} & $71.02\%$ & 1.57 MB \\
BGD~\cite{zeno2018task} & $80.44\%$ & 2.7 MB \\
TACOS~\cite{soures2021tacos} & $82.56\%$ & 1.5 MB \\
SS~\cite{schug2021presynaptic} & \textbf{$82.9\%$} & 1.57 MB \\
Online EWC~\cite{schwarz2018progress} & $64.34\%$ & 1.5 MB \\
\textbf{Genesis(Our work)} & $74.46\%$ & \textbf{628 kB} \\
\bottomrule
\end{tabular}
\label{tab:accuracy_tab}
\end{center}
\vspace{-3mm}

\end{table}
For evaluation, a spiking neural network with a single hidden layer containing 200 neurons and 2 output neurons is utilized. We conduct network-level simulations in python while considering the hardware constraints, such as low-preicsion quantized weights, dyadic constants, and bilinear metaplasticity functions, and to emulate in situ learning. To minimize training time, the network is trained on a reduced subset of the MNIST dataset, consisting of 10,000 training samples and 2,500 testing samples.


Figure~\ref{fig:Acc_results} presents the effects of quantization on the proposed continual learning model. The performance of the 16-bit fixed point (FXP) model is compared against the full precision floating point, 32-bit FXP, and a non continual learning model. While the full precision floating point model attains the mean accuracy of 79.69\%, we find an approximately 5\% degradation when parameters are quantized to 16-bit FXP (incorporated on Genesis) and 3\% when quantized to 32-bit FXP. The 16-bit FXP quantized model shows better capabilities in retaining the old knowledge by showing higher accuracy on task 1 but achieves lower accuracy on newer tasks compared to high precision models. 

When comparing our proposed model (metaplasticity-based approach) against other state-of-the-art regularization CL models, it is found that our approach outperforms the online EWC and LwF models by 11\% and 3\%, respectively (see Table~\ref{tab:accuracy_tab}). Although our model does not achieve high performances such as TACOS~\cite{soures2021tacos} and SS~\cite{schug2021presynaptic} techniques, it requires less than half the memory, making it suitable for edge devices with limited resources.


\subsection{Power consumption, Area, and Latency}
We evaluate the power consumption and area of the Genesis accelerator by synthesizing the design on an IBM 65nm CMOS technology node using the Synopsys design compiler (DC) tool. It is found that the Genesis accelerator consumes an average power of 17.1 mW while training on MNIST images scaled to 16$\times$16 pixels.~\fig{fig:poewr_area} depicts the breakdown of the power consumption and area of the accelerator by component in the presence and absence of metaplasticity. The power and area breakdown reveal that the PEs account for $55\%$ the on-chip area and $46\%$ power consumption. However, approximately $15\%$ of PE power consumption and $40\%$ of area is associated with metaplasticity, and this represents the incremental overhead incurred to support continual learning capabilities.

\begin{figure}[htbp]
\centering
\includegraphics[width=1.0\linewidth]{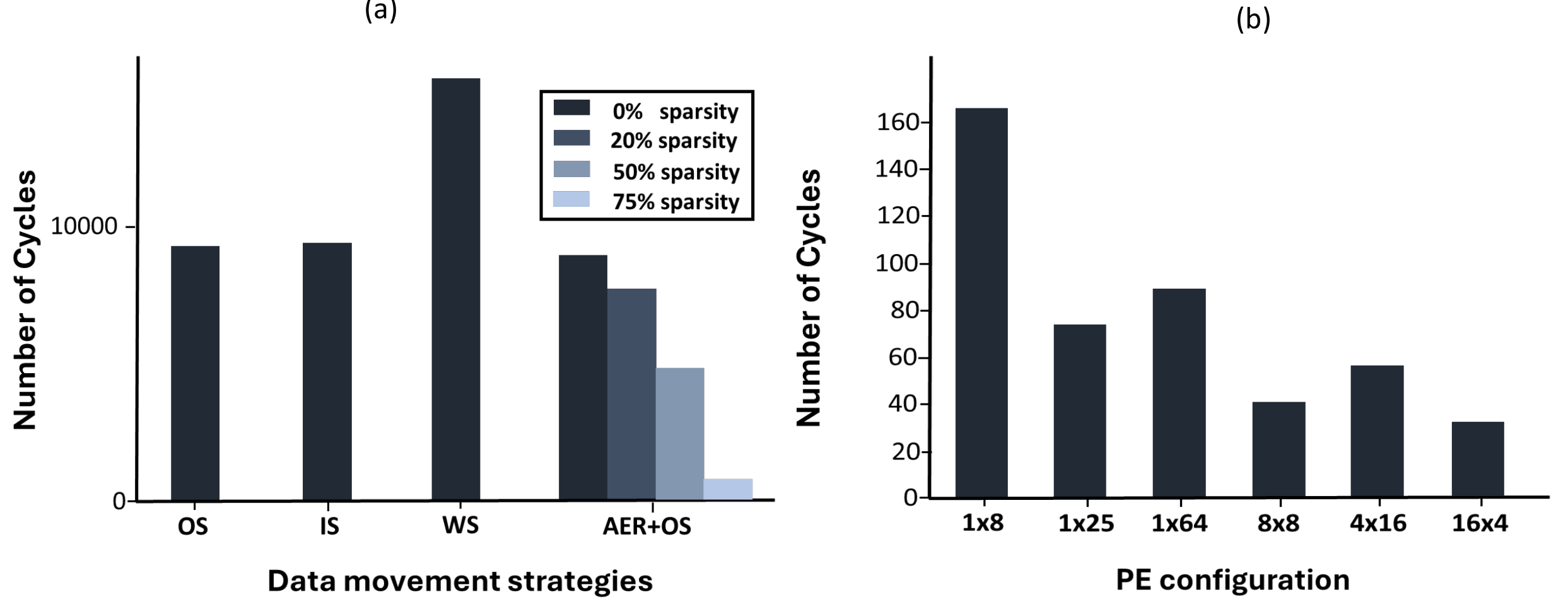}
      
\caption{(a) Latency of training spiking neural network with 256 input neurons ($25\%$ sparsity) and 256 output neurons while considering various strategies of data movements. (b) The impact of systolic array configurations on latency.}
\label{fig:latrncy_config} 
\end{figure}
\begin{figure}[!ht]
\centering
\includegraphics[width=1.0\linewidth]{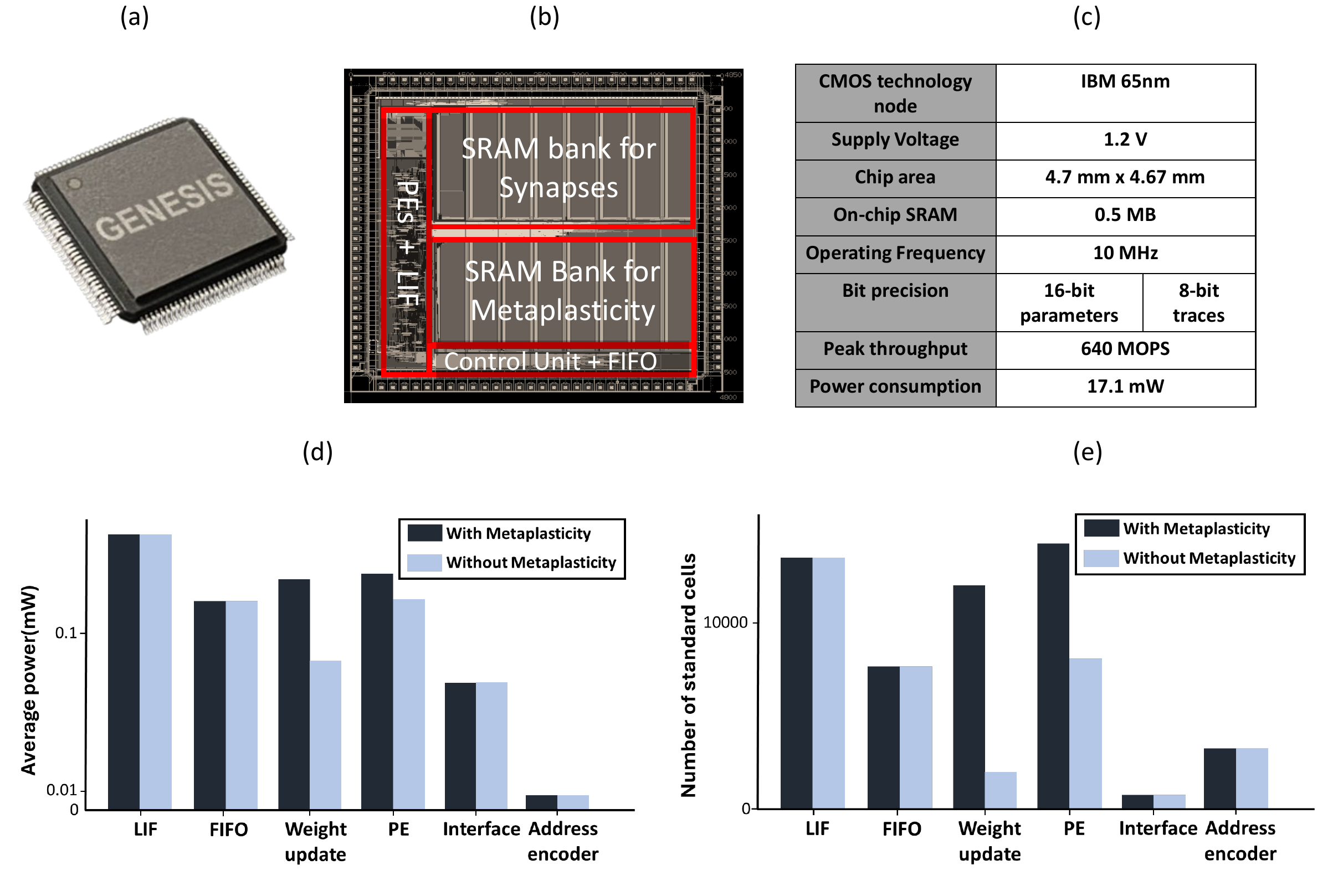}
      \caption{(a) Genesis chip that was fabricated on IBM65nm technology node. (b) The physical layout of the Genesis chip, annotating the memory for synapses and metaplasticity. (c) Specifications of the Genesis chip. (d) and (e) show the power and area breakdown for the different blocks and the potential overhead to incorporate metaplasticity.\vspace{-3mm}}
      \label{fig:poewr_area}   
\end{figure}

When it comes to latency, Genesis has the capability to train each MNIST image of 16x16 pixels in 10 milliseconds. This low latency can be attributed to the data movement strategy that was employed in processing;~\ref{fig:latrncy_config}(b). 
To identify the efficacy of the proposed strategy, we compare it against standard dataflows, such as output stationary (OS), weight stationary (WS), and input stationary (IS). For a spiking network with 256 input neurons and 256 output neurons simulated on the SCALEsim cycle simulator, it is observed that 
the proposed data movement strategy is $2.86\times$ and $1.08\times$ faster compared to WS and IS dataflows and performs similar to OS dataflow. When the inputs are sparse, the proposed strategy is $11.2\times$, $4.1\times$ and $4.25\times$ faster than IS, OS, and WS dataflows, respectively. 


\section{Conclusions}
We introduce Genesis, a spiking neuromorphic accelerator designed for continual learning, incorporating activity-dependent metaplasticity for robust knowledge retention while learning sequential tasks. The computational and memory overhead associated with continual learning is mitigated through two key contributions: low-precision 16-bit fixed-point quantization and a novel data movement strategy. The strategy with address event representation encoding to identify active neurons avoids unnecessary parameter movement and significantly reduces energy consumption. Genesis consumes only 17.08 mW while offering $74.46\%$ mean accuracy on the Split-MNIST setting for task-agnostic domain-incremental learning. We demonstrate that Genesis is ideal for resource-constrained environments by showing competitive performance and power with the state-of-the-art benchmarking for continual learning.



%

\bibliographystyle{IEEEtran}
\bibliography{ref.bib}

\end{document}